\documentclass[runningheads]{llncs}

\usepackage{eccv}

\usepackage{eccvabbrv}

\usepackage{graphicx}
\usepackage{booktabs}

\usepackage[accsupp]{axessibility}  

\usepackage[pagebackref,breaklinks,colorlinks,citecolor=eccvblue]{hyperref}

\usepackage{orcidlink}

\usepackage{multirow}
\usepackage{xcolor}
\usepackage{pifont}
\newcommand{\cmark}{\ding{51}} 
\newcommand{\xmark}{\ding{55}} 
\newcommand{\ours}{MPAD}

\makeatletter
\def\@fnsymbol#1{\ensuremath{\ifcase#1\or *\or \dagger\or \ddagger\or
   \mathsection\or \mathparagraph\or \|\or **\or \dagger\dagger
   \or \ddagger\ddagger \else\@ctrerr\fi}}
\makeatother

\begin{document}

\title{Unified Multi-plane Autoregressive Diffusion for 3D Multi-Contrast MRI Synthesis} 

\titlerunning{Unified Multi-Plane Autoregressive Diffusion}

\author{Yejee Shin\thanks{Equal contribution.}\inst{1}\and
Geonhui Son\inst{*}\thanks{Work done during internship at Microsoft Research Asia.}\inst{1,2}\and
Jinglu Wang\inst{2} \and
Minwoo Jung\inst{1} \and
Yan Lu\inst{\ddagger2} \and
Dosik Hwang\thanks{Corresponding authors.}\inst{1,3}}

\authorrunning{Y.~Shin et al.}

\institute{
Yonsei University \and
Microsoft Research Asia \and
Korea Institute of Science and Technology
}

\maketitle

\begin{abstract}
Acquiring a complete set of magnetic resonance imaging (MRI) contrasts is time-intensive and uncomfortable for patients, despite the diagnostic value of multi-contrast imaging. This motivates synthesizing missing contrasts from those already acquired, which is an inherently 3D problem requiring anatomical coherence across axial, sagittal, and coronal planes. However, fully 3D generative models are often impractical under computational resources that scale cubically with volume size.
We propose a unified Multi-Plane Autoregressive Diffusion (\ours), a latent diffusion framework that achieves full-volume 3D synthesis using efficient plane-wise 2D operations while preserving volumetric coherence. 
A 3D autoencoder first compresses MRI scans into an isotropic 3D latent representation. A 2D diffusion model is then trained to reconstruct masked latent slices of the target contrast, conditioned on both source-contrast slices and unmasked target-contrast slices.
During inference, we introduce plane-wise autoregressive synthesis with inter-plane priors. 
Slices are generated autoregressively in random order within one plane orientation to maintain intra-plane continuity, then propagated as conditioning priors to orthogonal plane orientations to enforce inter-plane consistency.
Compared to 3D latent diffusion baselines, \ours{} reduces training and inference FLOPs by 7× and 3×, respectively, while also lowering inference time and peak memory consumption.
Experiments on multiple datasets demonstrate that \ours{} achieves superior performance, generating high-fidelity 3D volumes and supporting one-to-many translation within a single unified model.
  \keywords{Multi-contrast MRI \and Synthesis \and Diffusion Model}
\end{abstract}

\section{Introduction}
\label{sec:intro}

\begin{figure}[!t]
    \centering
    \includegraphics[width=1.0\columnwidth]{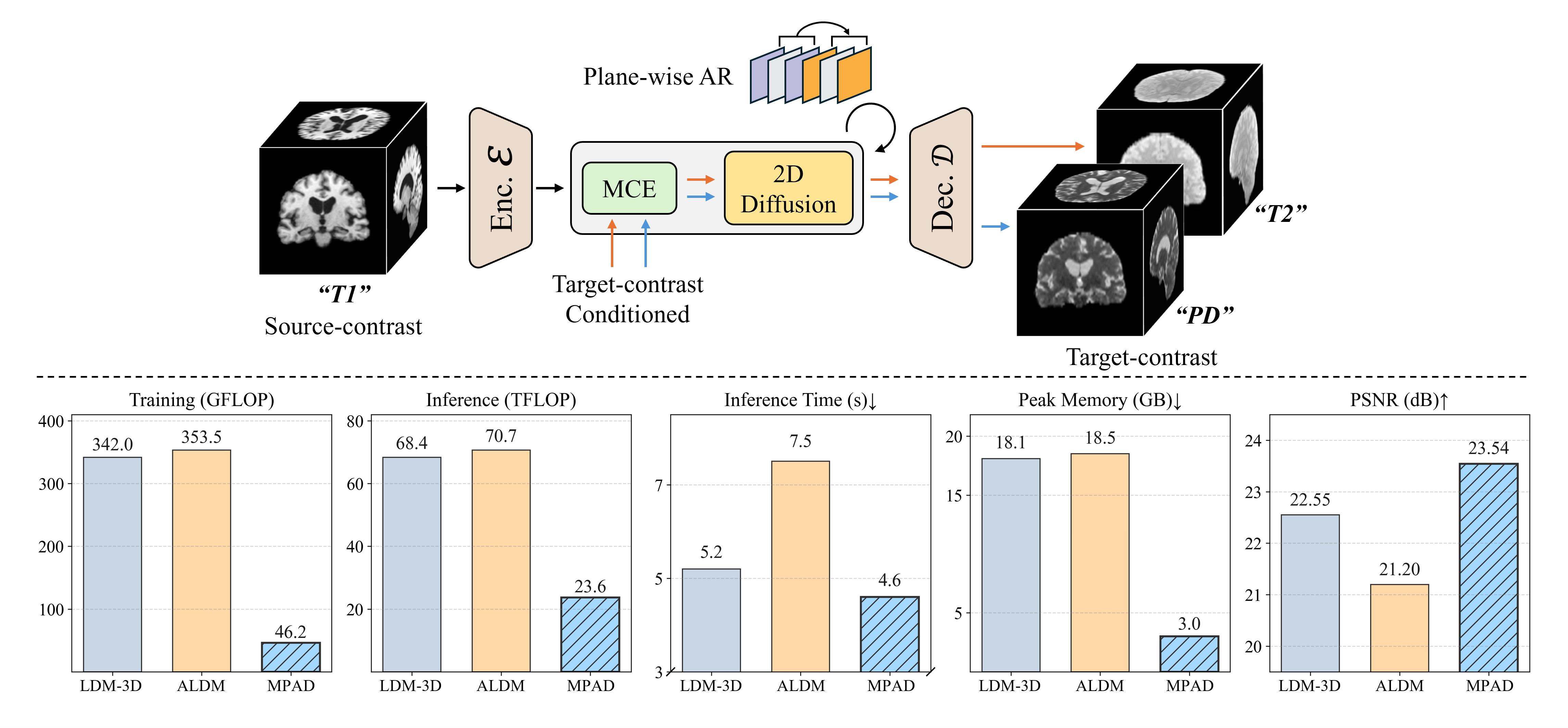}
\caption{Overview of the proposed 3D multi-contrast MR image translation framework. Given a source-contrast (e.g., T1), the model synthesizes a target contrast (e.g., T2 or PD) in latent space using a Multi-modal Conditioning Encoder (MCE) and a 2D diffusion model with plane-wise autoregressive synthesis. Bottom: Training FLOPs per iteration (GFLOPs), inference FLOPs per volume (TFLOPs), inference time, peak memory, and  PSNR averaged across all translation tasks on the ADNI dataset.}
\label{figure0}
\end{figure}

Magnetic resonance imaging (MRI) examinations routinely acquire multiple contrasts, each emphasizing different tissue characteristics and pathological cues. However, collecting all contrasts for every subject increases acquisition time, elevates patient discomfort, and burdens scanner throughput; prolonged scans also heighten susceptibility to motion artifacts that degrade image quality \cite{zaitsev2015motion,wood1985mr,lustig2007sparse}. 
The challenge is particularly severe for 3D volumetric acquisitions, which require substantially longer scan times due to their increased spatial coverage. This motivates the task of multi-contrast MR image synthesis, which aims to infer missing contrasts from acquired ones, enabling faster scans and more standardized imaging workflows.

Early work predominantly treated multi-contrast synthesis as a 2D image-to-image translation problem, learning mappings between contrasts such as T1-weighted to T2-weighted images. Generative adversarial networks (GANs) \cite{goodfellow2014generative} and variational autoencoders (VAEs) \cite{kingma2013auto} have been applied to convert slices between contrasts \cite{sharma2019missing, zhou2020hi,kong2021breaking,isola2017image,chen2020reusing,jang2025mint}. 
Although effective in 2D applications, these methods lack modeling of 3D volumetric coherence, resulting in inter-plane inconsistencies and anatomical distortions that hinder their direct application to 3D data.

For 3D multi-contrast MR image synthesis, recent studies improve comprehensive spatial continuity by employing fully volumetric 3D generative architectures, including 3D GAN and 3D diffusion models \cite{roca2025iguane,abramian2019generating}.
Although these models better capture anatomical structure, their computational and memory requirements grow cubically with volume size. This limits scalability, especially when a single model is expected to synthesize multiple target contrasts. 
Latent diffusion frameworks \cite{rombach2022high} partly mitigate this by performing generation in a compressed latent space rather than directly in the high-dimensional image space. 
Nevertheless, even with latent compression, 3D latent diffusion models face substantial practical constraints, including limited computational efficiency, prolonged inference time, and challenges in maintaining synthesis fidelity across multiple contrasts, all of which are critical considerations for real-world clinical applications.

In this paper, we aim to achieve efficient 3D multi-contrast MR image synthesis using a 2D diffusion model. We propose a unified multi-plane autoregressive diffusion (\ours) synthesis framework that enables efficient slice-wise generation while preserving 3D anatomical consistency.
First, we encode MR volumes into an isotropic latent representation with a 3D autoencoder, which allows for consistent slicing along any anatomical plane orientation.
Then, a 2D diffusion model is trained to reconstruct masked target-contrast slices conditioned on the source-contrast slices, unmasked target-contrast slices, and a text embedding specifying the desired target modality. 
By operating on 2D latent slices instead of full 3D volumes, our formulation avoids the cubic computational scaling of volumetric diffusion while reducing cost and preserving volumetric conditioning.

At inference, \ours~performs volumetric synthesis through a novel multi-plane autoregressive strategy that generates 2D slices across orthogonal planes, enforcing global 3D structural coherence.
Within each plane orientation, slices are generated autoregressively in random order to maintain intra-plane continuity. Crucially, when transitioning to an orthogonal plane orientation, the previously synthesized volume serves as a conditioning prior (inter-plane prior inference), progressively enriching the anatomical context and ensuring inter-plane consistency.

This plane-wise autoregressive strategy achieves substantial computational savings.
As shown in \cref{figure0}, \ours{} reduces training FLOPs by 7× and inference FLOPs by 3× compared to 3D latent diffusion baselines, while also significantly reducing inference time and peak memory consumption.
These improvements enable larger batch sizes, and practical deployment on standard GPUs.
Beyond these computational benefits, ~\ours{} also achieves superior reconstruction quality compared to 3D baselines.
In summary, our contributions are as follows:
\begin{itemize}
    \item We introduce a novel framework that reformulates 3D multi-contrast MR image synthesis as an efficient 2D masked-slice prediction task, overcoming the computational limits of 3D models.
    \item We propose a multi-plane autoregressive inference scheme that builds global 3D consistency from 2D operations using inter-plane priors.
    \item We demonstrate state-of-the-art efficiency and quality across multiple datasets, achieving 7× and 3× FLOP reductions in training and inference, significantly reduced inference time and peak memory, while improving synthesis fidelity.
    
\end{itemize}

\section{Related Works}
\label{sec:related_works}

\textbf{Multi-contrast MR image synthesis.}
Most previous studies in multi-contrast MR image synthesis have utilized 2D slice-based generative models, including GANs and diffusion models, to achieve strong fidelity in one-to-one or many-to-many translations. 
A representative many-to-many method is MM-GAN~\cite{sharma2019missing}, which synthesizes multiple missing sequences in a single forward pass.
Subsequently, diffusion-based methods like M2DN~\cite{meng2024multi} and APT~\cite{shin2025anatomical} were introduced to overcome typical GAN limitations such as mode collapse and limited diversity. However, when applied to 3D volume synthesis in a slice-by-slice manner, a fundamental limitation of these 2D approaches is their lack of explicit mechanisms to ensure volumetric coherence, which often results in anatomical inconsistencies between adjacent slices.

To address this, some works have explored full 3D volumetric synthesis. Methods in this category range from direct 3D adaptations of 2D architectures~\cite{mahboubisarighieh2024assessing} to more sophisticated models like cWDM~\cite{friedrich2024cwdm}, which applies diffusion to 3D wavelet coefficients.
In parallel, 2D/2.5D approaches attempt to approximate volumetric coherence by exchanging information between neighboring slices~\cite{cho2024two,hu2024diffgepci}. While these efforts confirm the importance of 3D consistency, they struggle to enforce long-range dependencies without incurring the prohibitive cost of true 3D generation. This creates a clear need for a framework that can achieve global 3D coherence efficiently, without relying on a full 3D generative model.

\noindent\textbf{Latent diffusion for 3D MRI.}
Recent advances in latent diffusion~\cite{rombach2022high} have enabled high-fidelity volumetric MR image generation. Generative models like Medical Diffusion~\cite{khader2023denoising} and BrainLDM~\cite{pinaya2022brain} learn a 3D generative prior over the latent space of medical images. Make-A-Volume~\cite{zhu2023make} restores 3D consistency by adding lightweight volumetric layers to a 2D slice-wise model, while the Volumetric Conditioning Module (VCM)~\cite{ahn2025volumetric} introduces a plug-in module for data-efficient control of pretrained 3D models.
Adaptive Latent Diffusion (ALDM)~\cite{kim2024adaptive} employs a 3D autoencoder with switchable SPADE~\cite{park2019semantic} blocks for multi-contrast conditioning.
While these methods validate the potential of latent diffusion for preserving volumetric context, their reliance on 3D denoising networks keeps computational requirements substantial.

\noindent\textbf{Autoregressive image generation.}
Autoregressive (AR) generative models predict image content sequentially by conditioning each output on previously generated ones. Early AR models such as PixelRNN and PixelCNN~\cite{van2016pixel, salimans2017pixelcnn++} operated directly in pixel space, and their sequential nature led to high computational cost.
Subsequent approaches shifted to token-based generation, where images are first discretized into latent tokens. Models like DALL-E~\cite{ramesh2021zero} and VQ-GAN~\cite{esser2021taming} demonstrated that AR prediction over learned latent codes yields higher fidelity with substantially reduced complexity. Efficiency was further improved with masked prediction and parallel decoding strategies, as seen in MaskGIT~\cite{chang2022maskgit} and multi-scale token prediction models such as VAR~\cite{tian2024visual}. More recent hybrid approaches introduce continuous token modeling, e.g., GIVT~\cite{tschannen2024givt} uses Gaussian mixture prediction, while MAR~\cite{li2024autoregressive} integrates diffusion-based denoising into the AR decoding loop.

We adapt the core principle of autoregression (AR) for 3D MR image synthesis by factorizing a volume into sequences of 2D slices. Our approach propagates anatomical context across orthogonal planes, enforcing global 3D coherence while retaining the efficiency of slice-wise 2D diffusion and avoiding the memory footprint of full 3D models.

\begin{figure*}[!t]
    \centering
    \includegraphics[width=1.0\textwidth]{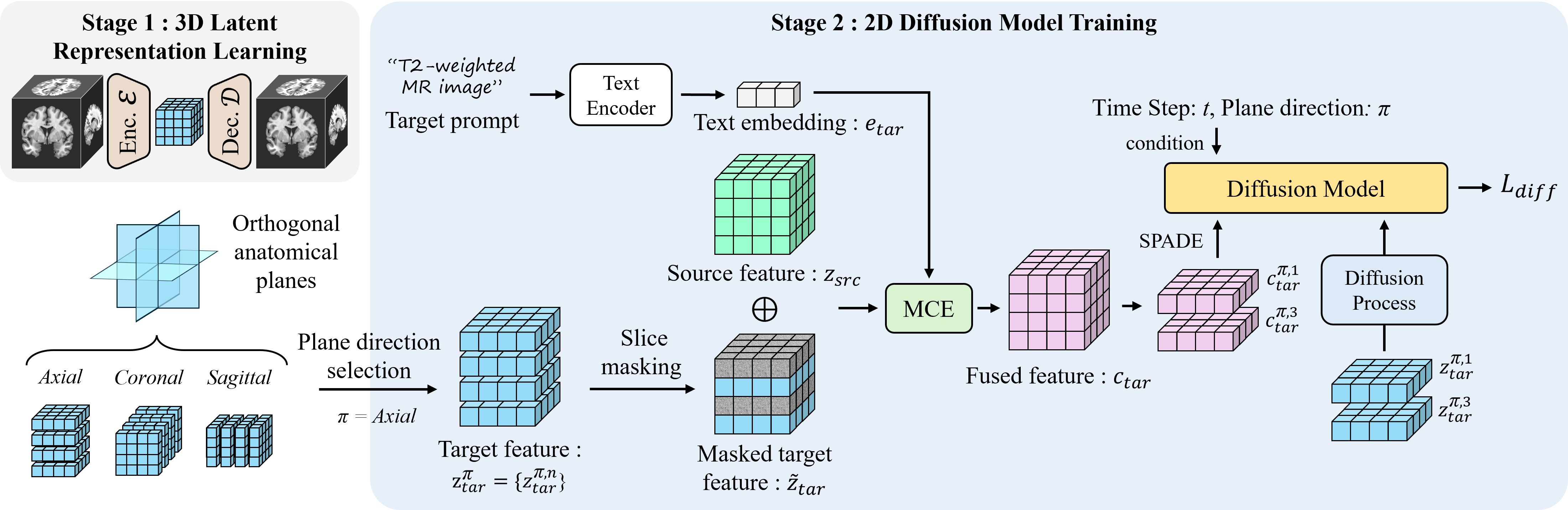}
\caption{Overview of the proposed training framework for 3D multi-contrast MR image synthesis. Stage 1: A 3D autoencoder encodes MR volumes into isotropic latent representations. Stage 2: A 2D diffusion model generates target-contrast latent slices conditioned on source-contrast slices, unmasked target-contrast slices, and text embeddings specifying the target modality. The Multi-modal Conditioning Encoder (MCE) fuses source and target features, which are then injected into the 2D diffusion model via SPADE modulation layers.
}
\label{figure1}
\end{figure*}

\section{Method}
We propose \ours, a framework for 3D multi-contrast MR image synthesis using a 2D autoregressive diffusion model operating in latent space. 
Training consists of two stages: (i) learning an isotropic 3D latent representation with a 3D autoencoder, and (ii) training a 2D latent diffusion model to reconstruct randomly masked target-contrast slices conditioned on source-contrast slices, both encoded in the 3D latent space.
An overview of the training framework is illustrated in \cref{figure1}.
At inference, we introduce a novel multi-plane autoregressive diffusion strategy.
\ours{} performs two coupled procedures: Intra-plane Autoregressive Synthesis (IAS) and Inter-plane Prior Inference (IPI).
IAS generates slices sequentially within each anatomical plane orientation to maintain intra-plane continuity, while IPI propagates previously generated volumes as inter-plane priors to enforce 3D structural coherence across orthogonal orientations. By combining these mechanisms, \ours{} achieves efficient slice-wise generation with global 3D consistency.

\subsection{3D Latent Representation Learning}
Let $X \in \mathbb{R}^{H \times W \times D \times C}$ denote a 3D MR image volume. 
Following LDM~\cite{rombach2022high}, we first train a KL-regularized autoencoder to compress data. The 3D autoencoder, parameterized by the encoder–decoder pair $(\mathcal{E}, \mathcal{D})$, maps an input volume to a latent tensor $z  \in \mathbb{R}^{h \times w \times d \times c}$ and reconstructs it from this latent representation:
\begin{equation}
\begin{aligned}
\min_{\mathcal{E}, \mathcal{D}} \; \mathcal{L}_{\text{ae}}(X, \hat{X})
\quad \text{s.t.} \quad
z = \mathcal{E}(X), \;
\hat{X} = \mathcal{D}(z),
\end{aligned}
\end{equation}
where the overall objective $\mathcal{L}_{\text{ae}}$ combines a reconstruction loss, a KL penalty on the approximate posterior, and an adversarial loss~\cite{goodfellow2014generative}.
This stage produces isotropic 3D latent features that support slicing along any spatial plane without geometric bias.

\subsection{2D Diffusion Training for Anatomical Planes}
Given a source contrast $X_{src}$ and target contrast $X_{tar}$, we extract their latent features as $z_{src} = \mathcal{E}(X_{src})$ and $z_{tar} = \mathcal{E}(X_{tar})$.
At each training iteration, a plane orientation $\pi \in \{\text{axial}, \text{coronal}, \text{sagittal}\}$ is sampled randomly. 
For each plane orientation $\pi$, the target latent $z_{tar}$ is sliced along $\pi$ and we denote the stacked slices as:
\begin{equation}
    z_{tar}=z_{tar}^{\pi}=\{z_{tar}^{\pi,n}\}_{n=1}^{N_s},
\end{equation}
where $N_s$ is the number of slices along that plane ($N_s=h,w,d$). 
A random subset of slice indices is masked to form a partially masked 3D latent $\tilde{z}_{tar}$. 

To inject target-contrast information, we encode the target prompt (e.g., “T2-weighted MR image”) with a pretrained multi-modal foundation model to obtain a text embedding denoted as $e_{\mathrm{tar}}$.
A 3D Multi-modal conditioning encoder (MCE) then produces conditioning features:
\begin{align}
    c_{\mathrm{tar}} &= \mathrm{MCE}(z_{\mathrm{src}}  \oplus \tilde{z}_{\mathrm{tar}},  e_{\mathrm{tar}}), \\
    c_{\mathrm{tar}} &= \{c_{\mathrm{tar}}^{\pi,n}\}_{n=1}^{N_s}.
\end{align}
The prompt is encoded by BioMedCLIP \cite{zhang2023biomedclip} to yield $e_{\mathrm{tar}}$, which provides semantic guidance for the desired output contrast \cite{byeon2025interpretable}.

To modulate intermediate 2D latent features with target-contrast information, we leverage spatially adaptive normalization (SPADE) \cite{park2019semantic} at each layer of the diffusion model. 
Given the 3D conditioning features $c_{\text{tar}}$ from MCE, we extract the corresponding 2D slice $c_{\text{tar}}^{\pi,n}$ aligned with the current slice position $n$ along the plane orientation $\pi$. 
The modulation of a latent feature map is defined as:
\begin{equation}
\begin{aligned}
    &\mathrm{SPADE}(z_\text{tar}^{\pi,n}, c_{\text{tar}}^{\pi,n}) \\ &= 
    \bigl(1 + \gamma(c_{\text{tar}}^{\pi,n})\bigr) \odot \mathrm{Norm}(z_\text{tar}^{\pi,n}) + \beta(c_{\text{tar}}^{\pi,n}),
\end{aligned}
\end{equation}
where $\gamma(c_{\text{tar}}^{\pi,n})$ and $\beta(c_{\text{tar}}^{\pi,n})$ are spatially varying modulation parameters inferred from the 2D conditioning slice, allowing the model to inject target-contrast cues at each spatial location. 
This layer-wise conditioning enables locally adaptive synthesis that aligns the generative process with the desired contrast.

The 2D diffusion model operates on individual latent slices. The model predicts target slices $z_{\mathrm{tar}}^{\pi,n}$ for masked positions, conditioned on the slice-level features $c_{\mathrm{tar}}^{\pi,n}$ and the plane identity $\pi$. Following the standard noise-prediction objective in diffusion models, we minimize
\begin{equation}
    \mathcal{L}_{\text{diff}} = \mathbb{E}_{\epsilon,\, t}\left[\,\|\epsilon - \epsilon_\theta(z_{tar,t}^{\pi,n}, t, c_{tar}^{\pi,n},\pi)\|_2^2\,\right],
\end{equation}
where $\epsilon_\theta$ denotes the denoising network, \(z_{tar,t}^{\pi,n}\) is the noisy latent at timestep \(t\), and $\epsilon$ is sampled from $\mathcal{N}(0, I)$.
During training, slices from different anatomical planes are jointly batched, so that the network is agnostic to the choice of plane orientation while capturing anatomical structure that is consistent across orientations.

\begin{figure*}[t]
    \centering
    \includegraphics[width=1.0\columnwidth]{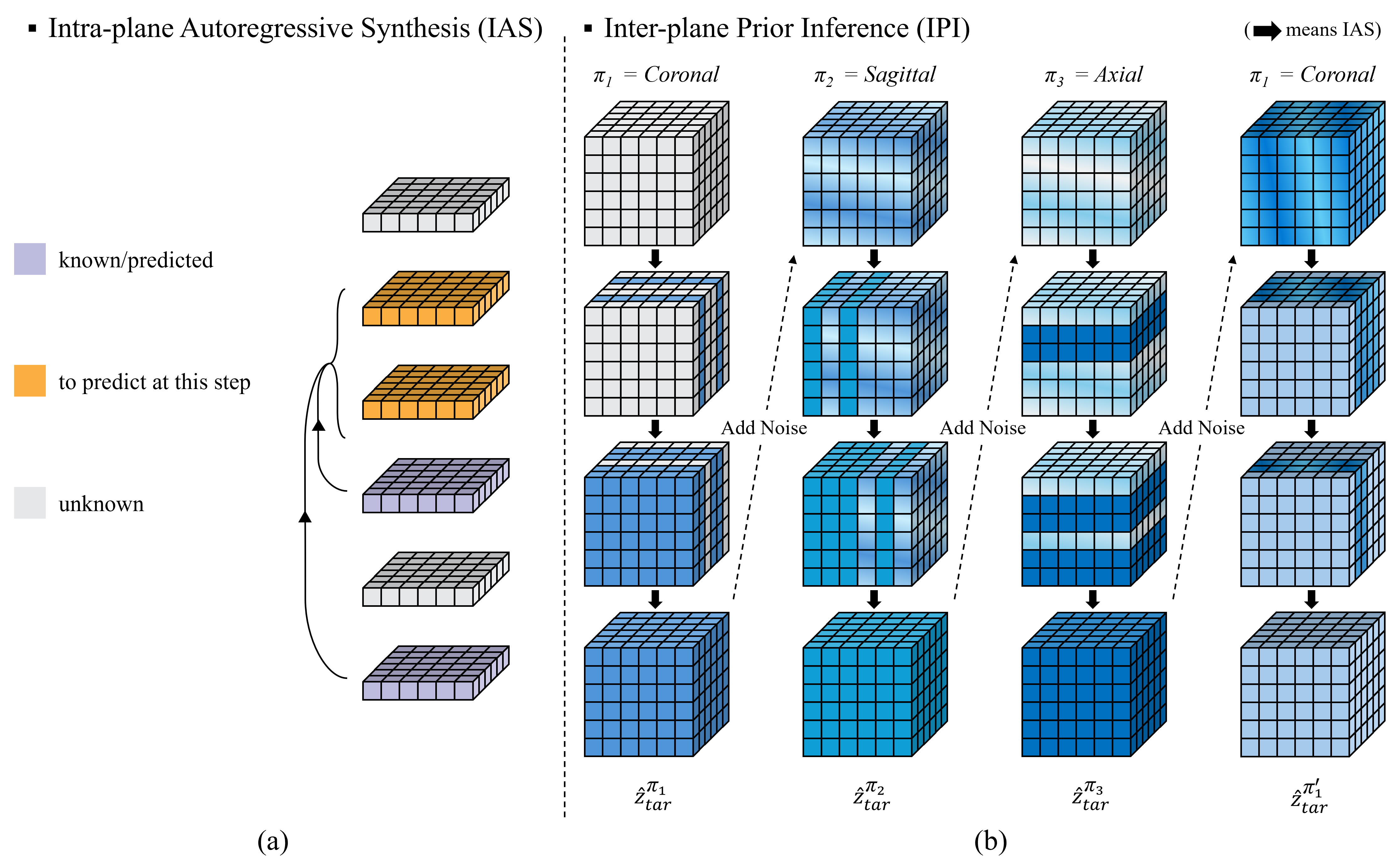}
\caption{(a): Random-order slice prediction within each plane, where slices are generated sequentially conditioned on previously predicted slices (purple) while unknown slices (gray) remain to be synthesized. (b): Four-pass generation process across planes. The first plane $\pi_1$ is generated from pure noise at timestep $T$. Subsequent planes $\pi_2$ and $\pi_3$ start from a smaller timestep $\tau$ using previously completed planes as priors. A refinement pass on $\pi_1$ incorporates information from all planes. The four candidates are aggregated via voxel-wise averaging to produce the final output.}
\label{inference}
\end{figure*}

\subsection{Plane-wise Autoregressive Synthesis}
At inference, we synthesize the target latent volume $\hat{z}_{tar}$ autoregressively along multiple plane orientations, as illustrated in \cref{inference}. The inference process operates through two coupled mechanisms. Intra-plane autoregressive synthesis (IAS) generates slices sequentially within each plane orientation to maintain spatial continuity. Inter-plane prior inference (IPI) leverages previously generated latent volumes as initialization priors for subsequent plane orientations, enforcing cross-plane consistency.

\noindent\textbf{Intra-plane Autoregressive Synthesis.}
Plane-wise synthesis along plane $\pi$ follows an autoregressive factorization. The conditional probability ${\textit{p}}\left(\hat{z}_{tar}^{\pi} \mid c_{tar}\right)$ takes the form:
\begin{equation}
{\textit{p}}\left(\hat{z}_{tar}^\pi \mid c_{tar}\right)=\prod_{n=1}^{N_s} {\textit{p}}\left(\hat{z}_{tar}^{\pi,n} \mid \hat{z}_{tar}^{\pi,<n}, c_{tar}\right),
\end{equation}
where $\hat{z}_{\mathrm{tar}}^{\pi,<n}$ denotes all previously generated slices.
The slice generation order is randomly sampled within each plane orientation.
The plane orientation order $(\pi_1, \pi_2, \pi_3)$ is defined as a permutation of $\{\text{coronal}, \text{sagittal}, \text{axial}\}$.

For the first plane orientation $\pi_1$, each slice is initialized from Gaussian noise and a reverse diffusion procedure is performed using the denoising network $\epsilon_\theta$ starting from the maximum timestep $T$. 
At each slice position $n$ and timestep $t$, all previously generated slices are encoded through MCE to produce a conditioning vector. The network then uses this conditioning, along with the timestep and plane identifier, to predict the denoised latent slice.
IAS ensures smooth transitions and anatomical continuity across slices through sequential conditioning within a plane.

\noindent\textbf{Inter-plane Prior Inference.} 
For subsequent planes $\pi_k$, IPI leverages previously completed latent volume as priors by initializing from a smaller intermediate timestep $\tau < T$ rather than pure noise. 
Specifically, for each slice position $n$ along plane $\pi_k$, we extract the corresponding slice from previously completed volumes and apply forward diffusion to reach timestep $\tau$. We then perform reverse diffusion from $\tau$ to $0$, conditioned on previously generated slices $\hat{z}^{\pi_k,<n}_{tar}$, source latents $z_{src}$, and target text embedding $e_{tar}$.
After completing all three planes, IPI performs a refinement pass on the first plane $\pi_1$, again starting from timestep $\tau$ by extracting slices from all three completed volumes. This produces a refined candidate $\hat{z}^{\pi_1'}_{tar}$ that incorporates information from all anatomical planes. By starting from a smaller timestep $\tau$ for planes with available priors, IPI reduces sampling steps while maintaining spatial consistency across the volume.
The four candidates $\{\hat{z}^{\pi_1}_{tar}, \hat{z}^{\pi_2}_{tar}, \hat{z}^{\pi_3}_{tar}, \hat{z}^{\pi_1'}_{tar}\}$ are aggregated using voxel-wise averaging:
\begin{equation}
\hat{z}_{tar} = \frac{1}{4}\left(\hat{z}^{\pi_1}_{tar} + \hat{z}^{\pi_2}_{tar} + \hat{z}^{\pi_3}_{tar} + \hat{z}^{\pi_1'}_{tar}\right).
\end{equation}
The final output volume is obtained by decoding the aggregated latent: $\hat{X}_{tar} = \mathcal{D}(\hat{z}_{tar})$. This multi-view aggregation ensures isotropic quality and reduces plane-specific artifacts by combining complementary information from all three anatomical orientations.

\section{Experiments and Analysis}

\subsection{Datasets}
We conduct experiments on two publicly available brain MRI datasets: ADNI~\cite{ADNI_dataset} and IXI~\cite{IXI_dataset}. ADNI consists of 737 MRI volumes from patients with Alzheimer's disease, acquired exclusively with 1.5T GE Medical Systems scanners. The protocol includes T1-weighted (T1w), T2-weighted (T2w), and proton density-weighted (PDw) sequences. IXI comprises 577 brain MRI volumes from healthy participants collected at different sites with varying acquisition parameters. The scans were obtained using 1.5T and 3T Philips scanners and include T1w, T2w, and PDw contrasts. 
For both ADNI and IXI, we randomly select 100 subjects for evaluation and use all remaining subjects for training. More details are provided in the supplementary material.

\subsection{Implementation Details}
We adopt a two-stage training strategy following the Latent Diffusion Model (LDM) \cite{rombach2022high} framework. 
In the first stage, we train a KL-regularized autoencoder to learn a compact latent representation of the 3D MRI volumes. The encoder compresses each volume into a latent space with $3$ channels at $32\times32\times32$ resolution, and the decoder reconstructs the original resolution volume. The autoencoder employs a 3D CNN architecture with channel multipliers of 1, 2, and 4, and two residual blocks per resolution level. We train the autoencoder for 1000 epochs.
In the second stage, we train a 2D CNN-based diffusion model in the learned latent space. The UNet backbone consists of 2 residual blocks per level and 32 channels per attention head. We employ a linear noise schedule ranging from $\beta_1 = 0.0015$ to $\beta_T = 0.0195$ over $T = 1000$ diffusion steps. During training, we apply random masking to the target latent slices with masking ratios uniformly sampled between 0.7 and 1.0. We use the AdamW optimizer with a learning rate of $4 \times 10^{-6}$ and train for 1000 epochs. To ensure a fair comparison, we train all competing methods for the same number of epochs. 

At inference, we perform autoregressive generation along three orthogonal planes using 10 DDIM steps. For the first plane, generation starts from random noise with the 10 timesteps. For subsequent planes, we leverage the previously generated volume as a prior and perform sampling using only 2 timesteps starting from an intermediate timestep $\tau$, thereby significantly reducing computational cost. 
We quantitatively evaluate synthesis performance using three metrics: peak signal-to-noise ratio (PSNR) (dB), structural similarity index (SSIM) \cite{wang2004image}, and normalized mean squared error (NMSE).

\subsection{Comparison Results}
We compare our method against five approaches for 3D MR image synthesis: two generative adversarial networks \cite{zhu2017unpaired,yu2019ea} and three diffusion-based methods \cite{rombach2022high,friedrich2024cwdm,kim2024adaptive}. Among these methods, only ALDM \cite{kim2024adaptive} has the one-to-many synthesis capability as ours, whereby a single model handles multiple target contrasts. The remaining baselines are trained separately for each contrast pair.

\Cref{tab:adni} and \Cref{tab:ixi} present quantitative comparisons on the ADNI and IXI datasets, respectively. Our method achieves competitive or superior performance across all contrast pairs and metrics. On the ADNI dataset, we obtain the highest SSIM scores in five out of six synthesis tasks and consistently achieve the lowest NMSE, demonstrating robust synthesis quality. 
Notably, for challenging tasks such as PD$\rightarrow$T1 and PD$\rightarrow$T2, our method significantly outperforms all baselines, improving PSNR by over 2 dB compared to the second-best method. 
On the IXI dataset, our approach consistently ranks first across all metrics, with particularly strong performance on T2$\rightarrow$T1 and PD$\rightarrow$T2 synthesis. These results demonstrate that our autoregressive framework achieves synthesis quality superior to 3D diffusion-based approaches while using only 2D denoising networks.

\begin{figure*}[!t]
    \centering
    \includegraphics[width=1.0\textwidth]{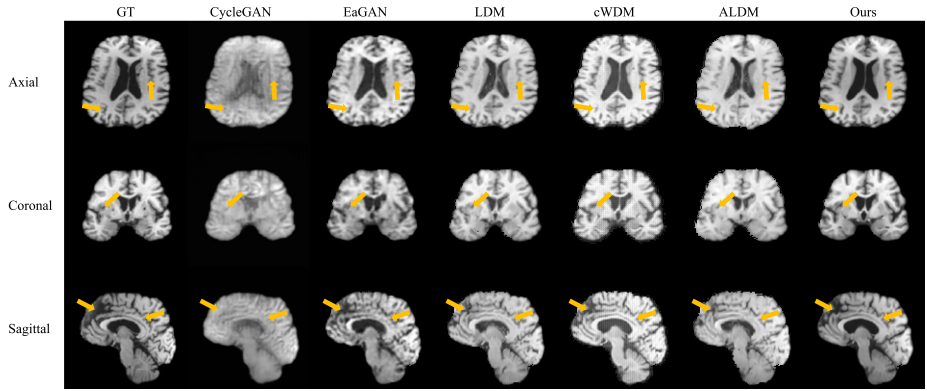}
\caption{Qualitative results on the ADNI dataset for T2→T1 synthesis. Each row shows axial, coronal, and sagittal views. Yellow arrows highlight anatomical regions where baseline methods show visible discrepancies compared to the ground truth (GT).}
\label{results_adni}
\end{figure*}

\begin{figure*}[!t]
    \centering
    \includegraphics[width=1.0\textwidth]{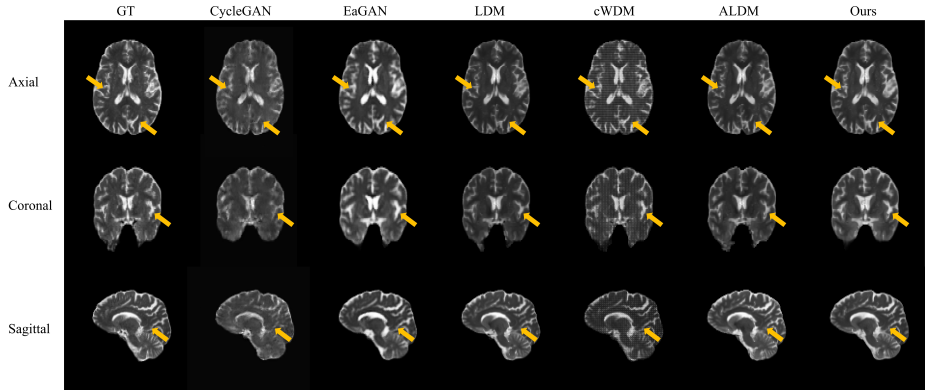}
\caption{Qualitative results on the IXI dataset for T1→T2 synthesis. Each row shows axial, coronal, and sagittal views. Yellow arrows highlight anatomical regions where baseline methods show visible discrepancies compared to the ground truth (GT).}
\label{results_ixi}
\end{figure*}

\cref{results_adni} and \cref{results_ixi} show qualitative comparisons on representative slices from both datasets. In \cref{results_adni}, we visualize T2$\rightarrow$T1 synthesis results across axial, coronal, and sagittal views. CycleGAN-3D and EaGAN produce blurring and loss of fine details, particularly in white matter boundaries and cortical structures. While LDM-3D and cWDM preserve overall anatomical structure, they introduce intensity distortions in some regions. ALDM shows improved structural fidelity but exhibits misalignments in complex anatomical regions. Our method produces sharper boundaries and more accurate tissue contrast across all three orthogonal planes. 
Similar tendencies are observed in \cref{results_ixi} for T1$\rightarrow$T2 synthesis on the IXI dataset, where our approach demonstrates better preservation of ventricular boundaries, gray-white matter contrast, and anatomical details compared to baselines. The plane-wise autoregressive generation maintains spatial coherence without introducing slice-wise artifacts.

\begin{table*}[!t]
\caption{Quantitative comparison on the ADNI dataset.  All metrics are evaluated on 3D volumes. Best scores are bolded.}
\centering
\resizebox{\textwidth}{!}{%
\begin{tabular}{ccccccccccccccccccc}
\hline
\multicolumn{1}{l}{} & \multicolumn{3}{c}{T1 $\rightarrow$ T2}           & \multicolumn{3}{c}{T1 $\rightarrow$ PD}           & \multicolumn{3}{c}{T2 $\rightarrow$ T1}           & \multicolumn{3}{c}{T2 $\rightarrow$ PD}           & \multicolumn{3}{c}{PD $\rightarrow$ T1}           & \multicolumn{3}{c}{PD $\rightarrow$ T2}           \\ \cline{2-19} 
Method               & PSNR$\uparrow$           & SSIM$\uparrow$           & NMSE$\downarrow$           & PSNR$\uparrow$           & SSIM$\uparrow$           & NMSE$\downarrow$           & PSNR$\uparrow$           & SSIM$\uparrow$           & NMSE$\downarrow$           & PSNR$\uparrow$           & SSIM$\uparrow$           & NMSE$\downarrow$           & PSNR$\uparrow$           & SSIM$\uparrow$           & NMSE$\downarrow$           & PSNR$\uparrow$           & SSIM$\uparrow$           & NMSE$\downarrow$           \\ \hline
CycleGAN-3D \cite{zhu2017unpaired}             & \textbf{24.88} & 0.810          & 0.130          & 20.09          & 0.774          & 0.179          & 21.49          & 0.807          & 0.197          & 19.63          & 0.761          & 0.204          & 19.08          & 0.751          & 0.248          & 19.23          & 0.751          & 0.450          \\
EaGAN \cite{yu2019ea}               & 21.72          & 0.826          & 0.273          & 19.80          & 0.815          & 0.203          & 21.14          & 0.829          & 0.199          & 20.32          & 0.834          & 0.184          & 19.54          & 0.779          & 0.269          & 20.42          & 0.784          & 0.362          \\
LDM-3D \cite{rombach2022high}                 & 22.89          & 0.803          & 0.214          & 23.59          & 0.816          & 0.085          & 21.63          & 0.817          & 0.161          & \textbf{25.29} & 0.843          & 0.068          & 19.82          & 0.772          & 0.261          & 22.07          & 0.785          & 0.257          \\
cWDM \cite{friedrich2024cwdm}                & 22.79          & 0.814          & 0.221          & 23.58          & 0.763          & 0.072          & 21.26          & 0.825          & 0.158          & 23.44          & 0.779          & 0.081          & 18.03          & 0.747          & 0.399          & 14.69          & 0.651          & 1.337          \\
ALDM \cite{kim2024adaptive}                & 20.77          & 0.770          & 0.352          & 22.96          & 0.797          & 0.099          & 20.15          & 0.789          & 0.236          & 24.00          & 0.816          & 0.085          & 19.15          & 0.754          & 0.307          & 20.15          & 0.751          & 0.412          \\
~\ours                 & 23.10          & \textbf{0.830} & \textbf{0.117} & \textbf{23.78} & \textbf{0.829} & \textbf{0.045} & \textbf{23.99} & \textbf{0.846} & \textbf{0.060} & 25.24          & \textbf{0.846} & \textbf{0.032} & \textbf{22.56} & \textbf{0.817} & \textbf{0.080} & \textbf{22.59} & \textbf{0.821} & \textbf{0.132} \\ \hline
\end{tabular}%
}

\label{tab:adni}
\end{table*}

\begin{table*}[!t]
\caption{Quantitative comparison on the IXI dataset. All metrics are evaluated on 3D volumes. Best scores are bolded.}
\centering
\resizebox{\textwidth}{!}{%
\begin{tabular}{ccccccccccccccccccc}
\hline
\multicolumn{1}{l}{} & \multicolumn{3}{c}{T1 $\rightarrow$ T2}           & \multicolumn{3}{c}{T1 $\rightarrow$ PD}           & \multicolumn{3}{c}{T2 $\rightarrow$ T1}           & \multicolumn{3}{c}{T2 $\rightarrow$ PD}           & \multicolumn{3}{c}{PD $\rightarrow$ T1}           & \multicolumn{3}{c}{PD $\rightarrow$ T2}           \\ \cline{2-19} 
Method               & PSNR$\uparrow$           & SSIM$\uparrow$           & NMSE$\downarrow$           & PSNR$\uparrow$           & SSIM$\uparrow$           & NMSE$\downarrow$           & PSNR$\uparrow$           & SSIM$\uparrow$           & NMSE$\downarrow$           & PSNR$\uparrow$           & SSIM$\uparrow$           & NMSE$\downarrow$           & PSNR$\uparrow$           & SSIM$\uparrow$           & NMSE$\downarrow$           & PSNR$\uparrow$           & SSIM$\uparrow$           & NMSE$\downarrow$           \\ \hline
CycleGAN-3D  \cite{zhu2017unpaired}           & 27.67 & 0.845          & 0.126          & 24.84          & 0.829          & 0.089          & 22.67          & 0.803          & 0.314          & 25.81          & 0.853          & 0.067          & 23.71          & 0.815          & 0.169          & 24.06          & 0.834          & 0.289          \\
EaGAN \cite{yu2019ea}               & 26.17          & 0.874          & 0.202          & 22.75          & 0.874          & 0.149          & 23.66          & 0.878          & 0.345          & 23.70          & 0.888          & 0.130          & 22.90          & 0.870          & 0.360          & 24.07          & 0.873          & 0.329          \\
LDM-3D \cite{rombach2022high}                  & 28.79          & 0.878          & 0.107          & 27.74          & 0.880          & 0.044          & 27.66          & 0.884          & 0.112          & 29.09 & 0.889          & 0.034          & 28.33          & 0.881          & 0.070          & 30.31          & 0.881          & 0.070          \\
cWDM \cite{friedrich2024cwdm}                & 24.62          & 0.853          & 0.319          & 27.28          & 0.870          & 0.057          & 20.86          & 0.836          & 0.526          & 26.26          & 0.862          & 0.086          & 24.53          & 0.850          & 0.235          & 21.98          & 0.836          & 0.585          \\
ALDM \cite{kim2024adaptive}              & 29.32          & 0.874          & 0.087          & 27.28          & 0.875          & 0.049          & 27.98          & 0.883          & 0.127          & 28.31          & 0.885          & 0.040          & 27.15          & 0.876          & 0.111          & 29.48          & 0.877          & 0.084          \\
\ours                 & \textbf{30.09} & \textbf{0.882} & \textbf{0.074} & \textbf{28.11} & \textbf{0.881} & \textbf{0.043} & \textbf{28.98} & \textbf{0.890} & \textbf{0.082} & \textbf{29.21} & \textbf{0.889} & \textbf{0.035} & \textbf{28.88} & \textbf{0.887} & \textbf{0.071} & \textbf{30.94} & \textbf{0.887} & \textbf{0.061} \\ \hline
\end{tabular}%
}

\label{tab:ixi}
\end{table*}

\subsection{Ablation Results}
\textbf{Effect of Autoregressive Synthesis.}
We evaluate IAS along a single anatomical plane by varying the group size $g$, which controls how many slices are generated in parallel at each autoregressive step (\Cref{tab:ar_ab}). Relative to the non-AR baseline, introducing IAS consistently improves reconstruction quality on both ADNI and IXI. Smaller group sizes yield incremental but steady gains, as more autoregressive steps enable finer slice-to-slice dependency modeling. 
Notably, even the non-AR baseline achieves competitive performance. This was driven by two factors. First, the source-contrast information $z_{src}$ is consistently provided for every slice, offering reliable anatomical guidance throughout the volume regardless of whether previous target slices are available. Second, since this source information is encoded through the 3D multi-modal conditioning encoder (MCE), it inherently captures volumetric structure and spatial context. This 3D-aware source conditioning ensures that the 2D diffusion model receives rich volumetric priors even without explicit autoregressive dependencies on previously generated target slices. However, autoregressive synthesis further improves upon this baseline by additionally providing previously generated target-contrast slices as conditioning, which explicitly enforces slice-to-slice continuity and refines local anatomical consistency in the target contrast. To balance accuracy and efficiency, we use $g{=}4$ in all experiments.

\begin{table}[!t]\small
\centering
\caption{Ablation study on the autoregressive (AR) framework. PSNR and NMSE are averaged across all six translation tasks (T1$\leftrightarrow$T2, T1$\leftrightarrow$PD, T2$\leftrightarrow$PD) on ADNI and IXI datasets.}
\begin{tabular}{cc|cccc}
\hline
   &         & \multicolumn{2}{c}{ADNI} & \multicolumn{2}{c}{IXI} \\
AR & $g$ & PSNR$\uparrow$        & NMSE$\downarrow$       & PSNR$\uparrow$       & NMSE$\downarrow$       \\ \hline
\xmark  & -       & 22.79       & 0.089      & 28.89      & 0.064      \\
\cmark  & 16       & 22.83       & 0.089      & 29.06      & 0.063      \\
\cmark  & 4       & 22.85       & 0.089      & 29.08      & 0.063      \\
\cmark  & 1      & 22.87       & 0.088      & 29.14      & 0.061      \\ \hline
\end{tabular}

\label{tab:ar_ab}
\end{table}

\begin{table}[!t]\small
\caption{Ablation study on multi-plane conditional encoding design. PSNR, SSIM, and NMSE are averaged across all six translation tasks (T1$\leftrightarrow$T2, T1$\leftrightarrow$PD, T2$\leftrightarrow$PD) on ADNI and IXI datasets. Each design choice is evaluated independently: MCE architecture (2D vs 3D convolution), masking type (learnable vs Gaussian noise), and prior information (off vs on). The first row in each block shows baseline performance, while the second row shows the performance change ($\Delta$) when switching the design. Bold values indicate the magnitude of improvement or degradation.}
\centering
\begin{tabular}{cc|cccccc}
\hline
                                 &                & \multicolumn{3}{c}{ADNI}                            & \multicolumn{3}{c}{IXI}                            \\
                                 &                & PSNR$\uparrow$   & SSIM$\uparrow$   & NMSE$\downarrow$  & PSNR$\uparrow$   & SSIM$\uparrow$  & NMSE$\downarrow$            \\ \hline
\multirow{2}{*}{MCE}             & 2D             & 22.82          & 0.813           & 0.091            & 28.18          & 0.879           & 0.090           \\
                                 & 3D             & \textbf{+0.72} & \textbf{+0.019} & \textbf{-0.013}  & \textbf{+1.19} & \textbf{+0.007} & \textbf{-0.029} \\ \hline
\multirow{2}{*}{Masking} & Learnable      & 23.22          & 0.803           & 0.133           & 27.37          & 0.869           & 0.102           \\
                                 & Gaussian Noise & \textbf{+0.32} & \textbf{+0.029} & \textbf{-0.055} & \textbf{+2.00} & \textbf{+0.017} & \textbf{-0.041} \\ \hline
\multirow{2}{*}{Prior}           & \xmark              & 22.42          & 0.795           & 0.087            & 27.49          & 0.873           & 0.098           \\
                                 & \cmark              & \textbf{+1.12} & \textbf{+0.037} & \textbf{-0.009}  & \textbf{+1.88} & \textbf{+0.013} & \textbf{-0.037} \\ \hline
\end{tabular}
\label{tab:ablation_3types}
\end{table}

\noindent\textbf{2D vs. 3D Multi-modal Conditioning Encoder.}
We compare 2D and 3D convolutional architectures in the multi-modal conditioning encoder (MCE). 
As shown in \Cref{tab:ablation_3types}, the 3D architecture consistently outperforms its 2D counterpart across both datasets.  
The 2D architecture processes each slice independently, so conditioning features are encoded without through-plane context, which limits its ability to capture fully 3D anatomical structures.
In contrast, 3D convolutions aggregate information across neighboring slices and depth, providing richer volumetric context and leading to more coherent volumes and stronger quantitative performance.

\noindent\textbf{Masking Strategy.}
We evaluate two masking approaches during training: learnable mask tokens versus Gaussian noise. \Cref{tab:ablation_3types} shows that Gaussian noise masking substantially outperforms learnable tokens on both datasets.
In our framework, the masked target latent is processed by MCE and then injected into the diffusion model via SPADE. 
When using a learned mask token, masked regions are replaced with constant values that MCE can easily detect and potentially ignore, reducing the usefulness of target-side context. 
In contrast, Gaussian noise produces continuous, spatially varying corruptions that remain within the natural range of latent activations, forcing MCE to extract robust features from the partially noised target latent.

\noindent\textbf{Effect of Inter-plane Priors.}
We assess the impact of incorporating inter-plane priors from previously generated planes during multi-plane inference. 
\Cref{tab:ablation_3types} demonstrates that using priors consistently improves performance.
Without priors, each plane is generated independently from pure noise and simply averaged, lacking cross-plane coordination. 
In contrast, incorporating inter-plane priors enables subsequent planes to leverage structural information from already completed planes, allowing the model to maintain anatomical consistency across orthogonal orientations. 
This use of inter-plane priors reduces the denoising burden by starting from a more informed initialization. It also enforces geometric constraints that prevent contradictory predictions between planes, ultimately enhancing both spatial coherence and overall image quality.

\begin{table}[!t]\small
\caption{Ablation study on multi-plane generation. PSNR, SSIM, and NMSE are averaged across all six translation tasks (T1$\leftrightarrow$T2, T1$\leftrightarrow$PD, T2$\leftrightarrow$PD) on ADNI and IXI datasets. Each row represents a different combination of planes used for generation. Checkmarks indicate which planes ($\pi_1$: coronal, $\pi_2$: sagittal, $\pi_3$: axial, $\pi_1'$: refined coronal) are generated and averaged. Best scores are bolded.}
\centering
\setlength\tabcolsep{3pt}
\begin{tabular}{cccc|cccccc}
\hline
\multicolumn{4}{c|}{Planes} & \multicolumn{3}{c}{ADNI} & \multicolumn{3}{c}{IXI} \\ \hline
$\pi_1$    & $\pi_2$   & $\pi_3$   & $\pi_{1}'$   & PSNR$\uparrow$   & SSIM$\uparrow$   & NMSE$\downarrow$  & PSNR$\uparrow$   & SSIM$\uparrow$  & NMSE$\downarrow$  \\ \hline
\cmark     &      &      &       & 22.82  & 0.820  & 0.089  & 29.17  & 0.883  & 0.060 \\
\cmark     & \cmark    &      &       & 23.32  & 0.828  & 0.081  & 29.00  & 0.884  & 0.067 \\
\cmark     &      & \cmark    &       & 23.27  & 0.827  & 0.081  & 29.25  & 0.884  & 0.065 \\
      & \cmark    & \cmark    &       & 23.33  & 0.827  & 0.079  & 29.15  & 0.885  & 0.064 \\
\cmark     & \cmark    & \cmark    &       & 23.49  & 0.830  & \textbf{0.078}  & 29.29  & \textbf{0.886}  & 0.062 \\
\cmark     & \cmark    & \cmark    & \cmark     & \textbf{23.54}  & \textbf{0.832}  & \textbf{0.078}  & \textbf{29.37}  & \textbf{0.886}  & \textbf{0.061} \\ \hline
\end{tabular}

\label{tab:ablation_planes}
\end{table}

\begin{table}[h]
\caption{Ablation study on the plane ordering for translation tasks (T1$\leftrightarrow$T2, T2$\leftrightarrow$PD, and PD$\leftrightarrow$T1) on the ADNI dataset.}
\centering
\small
\begin{tabular}{c|cc}
\hline
Planes Ordering    & PSNR $\uparrow$           & SSIM $\uparrow$           \\ \hline
$\pi_3$$\rightarrow$$\pi_1$$\rightarrow$$\pi_2$$\rightarrow$$\pi'_3$ & 23.48          & 0.830          \\
$\pi_2$$\rightarrow$$\pi_3$$\rightarrow$$\pi_1$$\rightarrow$$\pi'_2$ & 23.49          & 0.830          \\
$\pi_1$$\rightarrow$$\pi_2$$\rightarrow$$\pi_3$$\rightarrow$$\pi'_1$ & 23.54 & 0.832 \\ \hline
\end{tabular}
\label{tab:ablation_planes_ordering}
\end{table} 

\noindent\textbf{Effect of Multi-plane Generation.}
We analyze the effect of each anatomical plane and their combinations in our multi-plane generation framework. 
\Cref{tab:ablation_planes} shows a consistent trend of performance improvements as additional planes are incorporated. 
Starting from a single plane $\pi_1$ (coronal), adding a second plane $\pi_2$ (sagittal) already yields a noticeable gain in PSNR and SSIM on both ADNI and IXI, highlighting the benefit of complementary views. 
Generating along two orthogonal planes (e.g., $\pi_2$ and $\pi_3$) leads to similar improvements, indicating that the advantage comes from multi-view information rather than a particular choice of plane. 
Using all three anatomical planes ($\pi_1$, $\pi_2$, $\pi_3$) further boosts performance, and an additional refinement pass on the first plane ($\pi_1'$) after completing all three planes provides a small but consistent extra gain on both datasets. 
This refinement pass allows $\pi_1$ to leverage information from all completed views, reducing plane-specific artifacts and leading to more isotropic visual quality. 
Overall, these results demonstrate that multi-plane generation with a final refinement step effectively aggregates complementary spatial information across orientations, yielding more consistent and higher-quality 3D synthesis.

\noindent\textbf{Robustness of Plane Ordering.}
We analyze the robustness of plane ordering in the proposed IPI framework. As shown in \Cref{tab:ablation_planes_ordering}, different generation orders yield nearly identical performance across all evaluation metrics, indicating that the synthesis
process is largely invariant to the choice of plane sequence.
This robustness arises because each plane is generated while conditioned on both source-contrast information and previously synthesized volumes, allowing anatomical context to be progressively accumulated regardless of ordering. Consequently, inter-plane priors provide complementary structural guidance rather than order-dependent constraints, enabling stable multi-plane aggregation and consistent volumetric reconstruction.

\subsection{Discussion}
While our method substantially reduces inference cost compared to existing diffusion-based approaches (\cref{figure0}), the proposed multi-plane generation still involves sequential cross-plane refinement to enforce volumetric consistency. In our current implementation, reducing the number of inter-plane refinement iterations decreases inference time to approximately 1.8\,s per volume, indicating further efficiency headroom without architectural changes. This reflects a controllable trade-off between inference speed and multi-plane consistency. Future work could explore more efficient aggregation mechanisms or parallel generation strategies to further accelerate inference while maintaining the benefits of multi-view consistency. Additionally, distillation techniques or fewer-step sampling methods may offer promising directions for reducing the sampling burden without compromising synthesis quality.

\section{Conclusion}
We present \ours, a latent-space framework for 3D multi-contrast MRI synthesis that achieves full-volume generation using only 2D diffusion operations. A 3D autoencoder maps volumes to an isotropic latent representation that supports slicing along arbitrary plane orientations. The diffusion model is trained to reconstruct masked target-contrast slices conditioned on source-contrast slices and target modality specification. At inference, plane-wise autoregressive generation with inter-plane priors and multi-plane aggregation enforces volumetric consistency without full 3D diffusion operations.
By reformulating volumetric synthesis as coordinated 2D latent diffusion with structured cross-plane priors, \ours{} bridges computational efficiency and global 3D anatomical fidelity within a unified framework. Moreover, the controllable trade-off between inter-plane refinement and inference time highlights the practical flexibility of our approach, enabling adaptation to different deployment constraints. Future work includes further reducing sampling cost through distillation or parallelized multi-plane generation.

\section*{Acknowledgements}
This research was supported by the MSIT(Ministry of Science, ICT), Korea, under the Global Research Support Program in the Digital Field program(RS-2024-00436680) supervised by the IITP(Institute for Information \& Communications Technology Planning \& Evaluation), and Microsoft Research Asia.

\bibliographystyle{splncs04}
\bibliography{main}
\end{document}